\pdfoutput=1

\documentclass[11pt]{article}

\usepackage{naacl2021}

\usepackage{times}
\usepackage{latexsym}

\usepackage[T1]{fontenc}

\usepackage[utf8]{inputenc}

\usepackage{microtype}

\usepackage{amsmath}
\usepackage{amssymb}
\usepackage{interval}
\usepackage{subcaption}
\usepackage[nameinlink,capitalise]{cleveref}
\usepackage[inline]{enumitem}
\usepackage{booktabs}
\usepackage{tabularx}
\usepackage{multirow}
\usepackage{bm}
\usepackage{array}

\crefname{section}{\S}{\S\S}

\newcolumntype{x}[1]{>{\centering\arraybackslash\hspace{0pt}}p{#1}}

\usepackage{graphicx}
\graphicspath{{./img/}}

\newcommand{\para}[1]{\noindent\textbf{#1.}}

\newcommand{\meanvar}[2]{$#1$ {\tiny $\pm #2$}}

\newcommand{\tuple}[1]{\ensuremath{\left(#1\right)}}
\newcommand{\powerset}[1]{\ensuremath{\mathcal{P}\left(#1\right)}}
\newcommand{\eset}[1]{\left\{ #1 \right\}}
\newcommand{\cset}[2]{\left\{\, #1 \mid #2 \,\right\}}
\newcommand{\abs}[1]{\left|#1\right|}

\newcommand{\R}{\mathbb{R}}
\newcommand{\Z}{\mathbb{Z}}

\newcommand{\E}{L_E}  
\newcommand{\D}{L_D}  
\newcommand{\relset}{\mathcal{R}}
\newcommand{\entset}{\mathcal{E}}
\newcommand{\Dmax}{\delta_{\mathit{max}}}  
\newcommand{\hasType}{\textsc{has-type}}
\newcommand{\sameEnt}{\textsc{same}}
\newcommand{\same}{\ensuremath{\mathit{same}}}

\newcommand{\gtt}{KG$\to$text}
\newcommand{\softmax}[1]{\sigma\left(#1\right)}

\newcommand{\mat}{\bm}  

\renewcommand{\vec}{\bm}
\DeclareMathAlphabet{\mathsfit}{\encodingdefault}{\sfdefault}{m}{sl}
\SetMathAlphabet{\mathsfit}{bold}{\encodingdefault}{\sfdefault}{bx}{n}
\newcommand{\tens}[1]{\bm{\mathsfit{#1}}}

\title{Modeling Graph Structure via Relative Position for Text Generation from Knowledge Graphs}

\author{Martin Schmitt\textsuperscript{1}
	\, Leonardo F.\ R.\ Ribeiro\textsuperscript{2}
	\, Philipp Dufter\textsuperscript{1}
	\, Iryna Gurevych\textsuperscript{2}
	\, Hinrich Sch\"{u}tze\textsuperscript{1}\\\\
	\textsuperscript{1}Center for Information and Language Processing (CIS), LMU Munich\\
	\textsuperscript{2}Research Training Group AIPHES and UKP Lab, Technische Universität Darmstadt\\
	{\tt martin@cis.lmu.de}
}

\date{}

\begin{document}
\maketitle
\begin{abstract}
We present \emph{Graformer}, a novel Transformer-based encoder-decoder architecture for graph-to-text generation.
With our novel graph self-attention,
the encoding of a node relies on all nodes in the input graph
-- not only direct neighbors --
facilitating the detection of global patterns.
We represent
the relation between two nodes as
the length of the shortest path between them.
Graformer learns to weight these node-node relations
differently for different attention heads,
thus virtually learning differently connected views of the input graph.
We evaluate Graformer on two popular graph-to-text generation benchmarks,
AGENDA and WebNLG,
where it achieves strong performance
while using many fewer parameters
than other approaches.\footnote{Our code is publicly available: \url{https://github.com/mnschmit/graformer}}
\end{abstract}

\section{Introduction}
A knowledge graph (KG) is a flexible data structure
commonly used to store both general world knowledge \citep{auer08}
and specialized information, e.g., in biomedicine
\citep{wishart18}
and computer vision
\citep{visualgenome}.
Generating a natural language description of such a graph (\gtt{})
makes the stored information accessible to a broader audience of end users.
It is therefore important for KG-based question answering \citep{bhowmik-de-melo-2018-generating},
data-to-document generation \citep{moryossef-etal-2019-step,koncel-kedziorski-etal-2019-text}
and interpretability of KGs in general \citep{schmitt20}.

Recent approaches to \gtt{} employ encoder-decoder architectures:
the encoder first computes vector representations of the
graph's nodes, the decoder 
then uses them to predict the text sequence.
Typical encoder choices are graph neural networks
based on message passing between direct neighbors in the graph \citep{kipf17,velickovic18}
or variants of Transformer \citep{vaswani17}
that apply self-attention on all nodes together,
including those that
are not directly connected.
To avoid losing information,
the latter approaches use edge or node \textit{labels} from the
shortest path
when computing the attention between two nodes \citep{zhu-etal-2019-modeling,cai20}.
Assuming the existence of a path between any two nodes is
particularly problematic for KGs:
a set of KG facts often does not form a connected graph.

We propose a flexible alternative
that neither needs such an assumption nor uses label information to model graph structure:
a Transformer-based encoder
that interprets the \textit{lengths} of shortest paths in a graph as relative position information
and thus, by means of multi-head attention,
dynamically learns different structural views of the input graph with differently weighted connection patterns.
We call this new architecture \emph{Graformer}.

Following previous work,
we evaluate Graformer on two benchmarks:
\begin{enumerate*}[label={(\roman{*})}]
	\item the AGENDA dataset \citep{koncel-kedziorski-etal-2019-text}, i.e.,
	the generation of scientific abstracts
	from
	automatically extracted
	entities and relations specific to scientific text,
	and
	\item the WebNLG challenge dataset \citep{gardent-etal-2017-webnlg},
	i.e., the task of generating text from DBPedia subgraphs.
\end{enumerate*}
On both datasets,
Graformer achieves more than 96\%{} of the state-of-the-art performance
while using only about half as many parameters.

In summary, our contributions are as follows:
\begin{enumerate*}[label=(\arabic*)]
	\item We develop \emph{Graformer}, a novel graph-to-text architecture that interprets shortest path lengths as relative position information in a graph self-attention network.
	\item Graformer achieves competitive performance on two popular KG-to-text generation benchmarks, showing that our architecture can learn about graph structure without any guidance other than its text generation objective.
	\item To further investigate what Graformer learns about graph structure, we visualize the differently connected graph views it has learned and indeed find different attention heads for more local and more global graph information. Interestingly, direct neighbors are considered particularly important even without any structural bias, such as introduced by a graph neural network.
	\item Analyzing the performance w.r.t.\ different input graph properties,
	we find evidence that Graformer's more elaborate global view on the graph is an advantage
	when it is important to distinguish between distant but connected nodes and truly unreachable ones.
\end{enumerate*}

\section{Related Work}

Most recent approaches to graph-to-text generation
employ a graph neural network (GNN) based on message passing through the input graph's topology
as the encoder in their encoder-decoder architectures \citep{marcheggiani-perez-beltrachini-2018-deep,koncel-kedziorski-etal-2019-text,ribeiro-etal-2019-enhancing,guo-etal-2019-densely}.
As one layer of these encoders only considers immediate neighbors,
a large number of stacked layers can be necessary to learn about distant nodes,
which in turn also increases the risk of propagating noise \citep{li18}.

Other approaches \citep{zhu-etal-2019-modeling,cai20} base their encoder on the Transformer architecture \citep{vaswani17}
and thus, in each layer, compute self-attention on all nodes, not only direct neighbors,
facilitating the information flow between distant nodes.
Like Graformer, these approaches incorporate information about the graph topology with some variant of relative position embeddings \citep{shaw-etal-2018-self}.
They, however, assume that there is always a path between any pair of nodes, i.e., there are no unreachable nodes or disconnected subgraphs.
Thus they use an LSTM \citep{lstm} to compute a relation embedding from the
labels along this path.
However, in contrast to the AMR\footnote{abstract meaning representation} graphs used for their evaluation,
KGs are frequently disconnected.
Graformer is more flexible and makes no assumption about connectivity.
Furthermore, its relative position embeddings only depend on the \textit{lengths} of shortest paths
i.e., purely structural information, not labels.
It thus effectively learns \textit{differently connected views} of its input graph.

Deficiencies in modeling long-range dependencies in GNNs
have been considered a serious limitation before.
Various solutions orthogonal to our approach have been proposed in recent work:
By incorporating a connectivity score into their graph attention network,
\citet{zhang20} manage to increase the attention span to k-hop neighborhoods
but, finally, only experiment with $k=2$.
Our graph encoder efficiently handles dependencies between much more distant nodes.
\citet{pei20} define an additional neighborhood
based on Euclidean distance in a continuous node embedding space.
Similar to our work,
a node can thus receive information from distant nodes,
given their embeddings are close enough.
However, \citet{pei20} compute these embeddings only once before training
whereas in our approach
node similarity is based on the learned representation in each encoder layer.
This allows Graformer to dynamically change node interaction patterns during training.

Recently,
\citet{ribeiro20}
use two GNN encoders
-- one using the original topology and one with a fully connected version of the graph --
and combine their output in various ways for graph-to-text generation.
This approach
can only see two extreme versions of the graph: direct
neighbors and full connection.
Our approach is more flexible
and
dynamically learns a different structural view per attention head.
It is also more parameter-efficient as our multi-view encoder does not need a separate set of parameters for each view.

\section{The Graformer Model}

Graformer follows the general multi-layer encoder-decoder pattern
known from the original Transformer \citep{vaswani17}.
In the following,
we first describe our formalization of the KG input and then how it is processed by Graformer.

\begin{figure}[!ht]
	\begin{subfigure}{.9\linewidth}
		\centering
		\includegraphics[page=1,width=\linewidth]{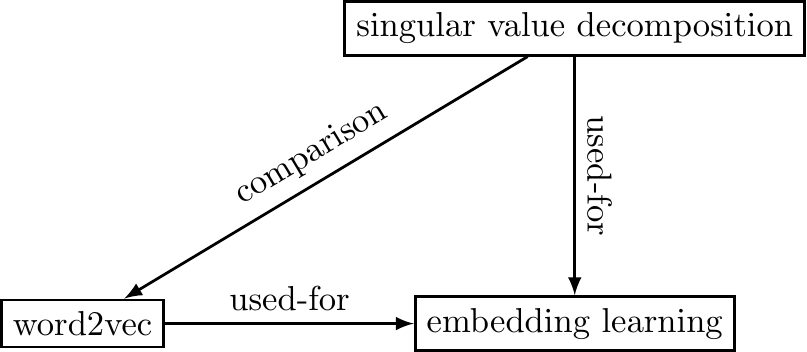}
		\subcaption{Original knowledge graph}
		\label{fig:graphs-1}
	\end{subfigure}
	
	\vspace{1em}
	\begin{subfigure}{\linewidth}
		\centering
		\includegraphics[page=2,width=\linewidth]{graphs}
		\subcaption{Directed hypergraph (token graph)}
		\label{fig:graphs-2}
	\end{subfigure}
	
	\vspace{.5em}
	\begin{subfigure}{\linewidth}
		\centering
		\includegraphics[page=3,width=\linewidth]{graphs}
		\subcaption{Incidence graph with $\sameEnt{}_p$ edges (dashed green)}
		\label{fig:graphs-3}
	\end{subfigure}
	\caption{Different representations of the same KG (types are omitted for clarity).}
	\label{fig:graphs}
\end{figure}

\subsection{Graph data structure}
\label{sec:data-structure}

\para{Knowledge graph}
We formalize a knowledge graph (KG) as a directed, labeled multigraph
$G_\mathit{KG} = \tuple{V, A, s, t, l_V, l_A, \entset, \relset}$
with $V$ a set of vertices (the KG entities),
$A$ a set of arcs (the KG facts), 
$s, t : A \to V$ functions assigning to each arc its source/target node (the subject/object of a KG fact),
and $l_V : V \to \entset, l_A : A \to \relset$ providing labels for vertices and arcs,
where $\relset$ is a set of KG-specific relations and $\entset$ a set of entity names.

\para{Token graph}
Entity names usually consist of more than one token or subword unit.
Hence, a tokenizer $\mathit{tok} : \entset \to \Sigma_T^*$ is
needed that
splits an entity's label into its components from the vocabulary $\Sigma_T$ of text tokens.
Following recent work \citep{ribeiro20},
we mimic this compositionality of node labels in the graph structure
by splitting each node into as many nodes as there are tokens in its label.
We thus obtain a directed hypergraph $G_T = \tuple{V_T, A, s_T, t_T, l_T, l_A, \Sigma_T, \relset, \same{}}$,
where $s_T, t_T : A \to \powerset{V_T}$ now assign a set of source (resp.\ target) nodes to each (hyper-) arc and all nodes are labeled with only one token,
i.e., $l_T : V_T \to \Sigma_T$.
Unlike \citet{ribeiro20},
we additionally keep track of all token nodes' origins:
$\same{} : V_T \to \powerset{V_T \times \Z}$ assigns to each node $n$ all other nodes $n'$ stemming from the same entity together with the relative position of $l_T(n)$ and $l_T(n')$ in the original tokenized entity name.
\cref{fig:graphs-2} shows the token graph corresponding  to the KG in \cref{fig:graphs-1}.

\para{Incidence graph}
For ease of implementation,
our final data structure for the KG is the hypergraph's incidence graph,
a bipartite graph where hyper-arcs are represented as nodes and edges are unlabeled:
$G = \tuple{N, E, l, \Sigma, \cset{\sameEnt{}_p}{p\in\Z}}$
where $N = V_T \cup A$ is the set of nodes,
$E = \cset{\tuple{n_1, n_2}}{n_1 \in s_T(n_2) \lor n_2 \in t_T(n_1)}$ the set of directed edges,
$l : N \to \Sigma$ a label function,
and $\Sigma = \Sigma_T \cup \relset$ the vocabulary.
We introduce $\sameEnt{}_p$ edges to fully connect \same{} clusters:
$\sameEnt{}_p = \cset{\tuple{n_1, n_2}}{\tuple{n_2, p} \in \same{}(n_1)}$
where $p$ differentiates between different relative positions in the original entity string,
similar to \citep{shaw-etal-2018-self}.
See \cref{fig:graphs-3} for an example.

\subsection{Graformer encoder}
\label{sec:graph-encoder}

The initial graph representation $\mat{H}^{(0)} \in \R^{\abs{N}\times d}$ is obtained by looking up embeddings for the node labels in the learned embedding matrix $\mat{E}\in \R^{\abs{\Sigma}\times d}$, i.e.,
$\vec{H}_i^{(0)} = \vec{e}^{l(n_i)} \mat{E}$ where $\vec{e}^{l(n_i)}$ is the one-hot-encoding of the $i$th node's label.

To compute the node representation $\mat{H}^{(L)}$ in the $L$th layer,
we follow \citet{vaswani17}, i.e.,
we first normalize the input from the previous layer $\mat{H}^{(L-1)}$ via layer normalization $\mathit{LN}$,
followed by multi-head graph self-attention $\mathit{SelfAtt}_g$ (see \cref{sec:self-att} for details),
which
-- after dropout regularization $\mathit{Dr}$ and a residual connection --
yields the intermediate representation $\mat{\mathcal{I}}$ (cf.\ \cref{eq:enc_inter}).
A feedforward layer $\mathit{FF}$ with one hidden layer and GeLU \citep{hendrycks16} activation
computes the final layer output (cf.\ \cref{eq:enc_final}).
As recommended by \citet{chen-etal-2018-best}, we apply an additional layer normalization step to the output $\mat{H}^{(\E)}$ of the last encoder layer $\E$.
\begin{align}
\label{eq:enc_inter}
\mat{\mathcal{I}}^{(L)} &= \mathit{Dr}(\mathit{SelfAtt}_g(\mathit{LN}(\mat{H}^{(L-1)}))) + \mat{H}^{(L-1)} \\
\label{eq:enc_final}
\mat{H}^{(L)} &= \mathit{Dr}(\mathit{FF}(\mathit{LN}(\mat{\mathcal{I}}^{(L)}))) + \mat{\mathcal{I}}^{(L)}
\end{align}
$\mathit{SelfAtt}_g$ computes a weighted sum of $\mat{H}^{(L-1)}$:
\begin{equation}
\label{eq:6}
\mathit{SelfAtt}_g(\mat{H})_i = \sum_{j=1}^{\abs{N}} \alpha_{ij}^g (\vec{H}_j \mat{W}^{V_{g}})
\end{equation}
where $\mat{W}^{V_g}\in\R^{d\times d}$ is a learned parameter matrix.

In the next section, we derive the definition of the graph-structure-informed attention weights $\alpha_{ij}^g$.

\subsection{Self-attention for text and graphs with relative position embeddings}
\label{sec:self-att}

In this section, we describe the computation of attention weights for multi-head self-attention.
Note that the formulas describe the computations for one head.
The output of multiple heads is combined as in the original Transformer \citep{vaswani17}.

\para{Text self-attention}
\citet{shaw-etal-2018-self} introduced position-aware self-attention in the Transformer
by
\begin{enumerate*}[label={(\roman*)}]
  \item adding a relative position embedding $\tens{A}^K \in \R^{M\times M\times d}$ to $\mat{X}$'s key representation, when computing the softmax-normalized attention scores $\vec{\alpha}_{i}$ between $\vec{X}_i\in\R^d$ and the complete input embedding matrix $\mat{X}\in\R^{M\times d}$ (cf.\ \cref{eq:1}), and
  \item adding a second type of position embedding $\tens{A}^V\in\R^{M\times M\times d}$ to $\mat{X}$'s value representation when computing the weighted sum (cf.\ \cref{eq:2}):
\end{enumerate*}
\begin{align}
  \label{eq:1}
  \vec{\alpha}_{i} &= \softmax{\frac{\vec{X}_i\mat{W}^Q (\mat{X}\mat{W}^K + \tens{A}^K_{i})^\top}{\sqrt{d}}} \\
  \label{eq:2}
  \vec{V}_i &= \sum_{j=1}^n \alpha_{ij} (\vec{X}_j\mat{W}^V + \tens{A}_{ij}^V)
\end{align}
where $\softmax{\cdot}$ denotes the softmax function, i.e.,
\begin{displaymath}
	\softmax{\vec{b}}_i = \frac{\exp\left(b_i\right)}{\sum_{j=1}^{J} \exp\left(b_j\right)},\quad \text{for $\vec{b}\in\R^{J}$.}
\end{displaymath}

Recent work \citep{raffel19} has adopted a simplified form
where value-modifying embeddings $\tens{A}^V$ are omitted
and key-modifying embeddings $\tens{A}^K$ are replaced
with learned scalar embeddings $\mat{S} \in \R^{M\times M}$ that -- based on relative position -- directly in- or decrease attention scores before normalization,
i.e., \cref{eq:1} becomes \cref{eq:3}.
\begin{equation}
  \label{eq:3}
  \vec{\alpha}_{i} = \softmax{\frac{\vec{X}_i\mat{W}^Q (\mat{X}\mat{W}^K)^\top}{\sqrt{d}} + \vec{S}_{i}}
\end{equation}

\citet{shaw-etal-2018-self} share their position embeddings across attention heads
but learn separate embeddings for each layer
as word representations from different layers can vary a lot.
\citet{raffel19} learn separate $\mat{S}$ matrices for each attention head but share them across layers.
We use \citet{raffel19}'s form of relative position encoding for text self-attention in our decoder (\cref{sec:decoder}).

\begin{figure}
  \small
	\centering
	\def\intercolumn{\hspace{.4em}}
	\def\outspace{\hspace{.4em}}
	\def\cellwidth{1.25em}
	\begingroup
	\renewcommand*{\arraystretch}{1.1}
	\begin{tabular}{@{\outspace}x{\cellwidth}@{\intercolumn}|@{\intercolumn}x{\cellwidth}@{\intercolumn}|@{\intercolumn}x{\cellwidth}@{\intercolumn}|@{\intercolumn}x{\cellwidth}@{\intercolumn}|@{\intercolumn}x{\cellwidth}@{\intercolumn}|@{\intercolumn}x{\cellwidth}@{\intercolumn}|@{\intercolumn}x{\cellwidth}@{\intercolumn}|@{\intercolumn}x{\cellwidth}@{\intercolumn}|@{\intercolumn}x{\cellwidth}@{\intercolumn}|@{\intercolumn}x{\cellwidth}@{\outspace}}
		&\multicolumn{6}{@{\intercolumn}c@{\intercolumn}|@{\intercolumn}}{$V_T$} & \multicolumn{3}{@{\intercolumn}c@{\intercolumn}}{$A$}\\[.25em]
		\hline
		& \texttt{s} & \texttt{v} & \texttt{d} & \texttt{w} & \texttt{e} & \texttt{l} & \texttt{c} & \texttt{u1} & \texttt{u2} \\
		\hline
		\texttt{s} & 0 & 4 & 5 & 2 & 2 & 2 & 1 & 1 & 3 \\
		\hline
		\texttt{v} & -4 & 0 & 4 & 2 & 2 & 2 & 1 & 1 & 3 \\
		\hline
		\texttt{d} & -5 & -4 & 0 & 2 & 2 & 2 & 1 & 1 & 3 \\
		\hline
		\texttt{w} & -2 & -2 & -2 & 0 & 2 & 2 & -1 & $\infty$ & 1 \\
		\hline
		\texttt{e} & -2 & -2 & -2 & -2 & 0 & 4 & -3 & -1 & -1 \\
		\hline
		\texttt{l} & -2 & -2 & -2 & -2 & -4 & 0 & -3 & -1 & -1 \\
		\hline
		\texttt{c} & -1 & -1 & -1 & 1 & 3 & 3 & 0 & $\infty$ & 2 \\
		\hline
		\texttt{u1} & -1 & -1 & -1 & $\infty$ & 1 & 1 & $\infty$ & 0 & $\infty$ \\
		\hline
		\texttt{u2} & -3 & -3 & -3 & -1 & 1 & 1 & -2 & $\infty$ & 0 \\
		\hline
	\end{tabular}
\endgroup
	\caption{$\mat{R}$ matrix for the graph in \cref{fig:graphs-3} ($\Dmax = 3$).}
	\label{fig:r_matrix}
\end{figure}

\para{Graph self-attention}
Analogously to self-attention on text,
we define our structural graph self-attention as follows:
\begin{equation}
  \label{eq:7}
  \vec{\alpha}_{i}^g = \softmax{\frac{\vec{H}_i\mat{W}^{Q_g} (\vec{H}\mat{W}^{K_g})^\top}{\sqrt{d}} + \gamma(\mat{R})_{i}}
\end{equation}
$\mat{W}^{K_g}, \mat{W}^{Q_g} \in \R^{d\times d}$ are learned matrices
and $\gamma : \Z \cup \eset{\infty} \to \R$ looks up \emph{learned scalar embeddings}
for the relative graph positions in $\mat{R}\in\R^{N\times N}$.

We define the relative graph position $R_{ij}$ between the nodes $n_i$ and $n_j$ with respect to two factors:
\begin{enumerate*}[label={(\roman{*})}]
	\item the text relative position $p$ in the original entity name
		if $n_i$ and $n_j$ stem from the same original entity, i.e., $\tuple{n_i, n_j}\in \sameEnt{}_p$ for some $p$
		and
	\item shortest path lengths otherwise:
\end{enumerate*}
\begin{equation}
  \label{eq:9}
  R_{ij} =
  \begin{cases}
	\infty, & \text{if } \delta(n_i, n_j) = \infty \\
	   & \text{and } \delta(n_j, n_i) = \infty \\
	\mathit{encode}(p), & \text{if } \tuple{n_i, n_j} \in \sameEnt{}_p \\
    \delta(n_i, n_j),   & \text{if } \delta(n_i, n_j) \leq \delta(n_j, n_i) \\
    -\delta(n_j, n_i),  & \text{if } \delta(n_i, n_j) > \delta(n_j, n_i)
  \end{cases}
\end{equation}
where $\delta(n_i, n_j)$ is the length of the shortest path from $n_i$ to $n_j$,
which we define to be $\infty$ if and only if there is no such path.
$\mathit{encode}$ maps a text relative position $p\in\Z\setminus\eset{0}$
to an integer outside $\delta$'s range to avoid clashes.
Concretely, we use $\mathit{encode}(p) := \mathit{sgn}(p)\cdot \Dmax + p$
where $\Dmax$ is the maximum graph diameter,
i.e., the maximum value of $\delta$ over all graphs under consideration.

Thus, we model graph relative position
as the length of the shortest path
using either only forward edges ($R_{ij} > 0$) or only backward edges ($R_{ij} < 0$).
Additionally, two special cases are considered:
\begin{enumerate*}[label={(\roman*)}]
\item Nodes without any purely forward or purely backward path between them ($R_{ij} = \infty$) and
\item token nodes from the same entity. Here the relative position in the original entity string $p$ is encoded outside the range of path length encodings (which are always in the interval $\interval{-\Dmax}{\Dmax}$).
\end{enumerate*}

In practice,
we use two thresholds, $n_\delta$ and $n_p$.
All values of $\delta$ exceeding $n_\delta$ are set to $n_\delta$
and analogously for $p$.
This limits the number of different positions a model can distinguish.

\para{Intuition}
Our definition of relative position in graphs combines several advantages:
\begin{enumerate*}[label=(\roman*)]
	\item Any node can attend to any other node -- even
          unreachable ones --
	while learning a suitable attention bias for different distances.
	\item $\sameEnt{}_p$ edges are treated differently in the attention mechanism.
	Thus, entity representations can be learned like in a regular transformer encoder,
	given that tokens from the same entity are fully connected with $\sameEnt{}_p$ edges
	with $p$ providing relative position information.
	\item The lengths of shortest paths
	often have an intuitively useful interpretation in our incidence graphs
	and the sign of the entries in $\mat{R}$ also captures
	the important distinction between incoming and outgoing paths.
	In this way, Graformer can, e.g.,
	capture the difference between the subject and object of a fact,
	which is expressed as a relative position of $-1$ vs.\ $1$.
	The subject and object nodes, in turn, see each other as $2$ and $-2$, respectively.
\end{enumerate*}

\cref{fig:r_matrix} shows the $\mat{R}$ matrix corresponding to the graph from \cref{fig:graphs-3}.
Note how token nodes from the same entity, e.g., \texttt{s}, \texttt{v}, and \texttt{d}, form clusters as they have the same distances to other nodes,
and how the relations inside such a cluster are encoded outside the interval $\interval{-3}{3}$,
i.e., the range of shortest path lengths.
It is also insightful to compare node pairs with the same value in $\mat{R}$.
E.g., both \texttt{s} and \texttt{w} see \texttt{e} at a distance of $2$
because the entities \emph{SVD} and \emph{word2vec} are both the subject of a fact with \emph{embedding learning} as the object.
Likewise, \texttt{s} sees both \texttt{c} and \texttt{u1} at a distance of $1$ because its entity \emph{SVD} is subject to both corresponding facts.

\subsection{Graformer decoder}
\label{sec:decoder}

Our decoder follows closely the standard Transformer decoder \citep{vaswani17},
except for the modifications suggested by \citet{chen-etal-2018-best}.

\para{Hidden decoder representation}
The initial decoder representation $\mat{Z}^{(0)} \in\R^{M\times d}$ embeds the (partially generated) target text $\mat{T}\in\R^{M\times\abs{\Sigma}}$,
i.e., $\mat{Z}^{(0)} = \mat{T} \mat{E}$.
A decoder layer $L$ then obtains a contextualized representation via self-attention as in the encoder (\cref{sec:graph-encoder}):
\begin{equation}
	\mat{C}^{(L)} = \mathit{Dr}(\mathit{SelfAtt}_t(\mathit{LN}(\mat{Z}^{(L-1)}))) + \mat{Z}^{(L-1)}
\end{equation}
$\mathit{SelfAtt}_t$ differs from $\mathit{SelfAtt}_g$ by using different position embeddings in \cref{eq:7}
and, obviously, $R_{ij}$ is defined in the usual way for text.
$\mat{C}^{(L)}$ is then modified via multi-head attention $\mathit{MHA}$ on the output $\mat{H}^{(\E)}$ of the last graph encoder layer $\E$.
As in \cref{sec:graph-encoder},
we make use of residual connections, layer normalization $\mathit{LN}$, and dropout $\mathit{Dr}$:
\begin{align}
	\mat{U}^{(L)} &= \mathit{Dr}(\mathit{MHA}(\mathit{LN}(\mat{C}^{(L)}), \mat{H}^{(\E)})) + \mat{C}^{(L)} \\
	\mat{Z}^{(L)} &= \mathit{Dr}(\mathit{FF}(\mathit{LN}(\mat{U}^{(L)}))) + \mat{U}^{(L)}
\end{align}
where
\begin{align}
	\label{eq:mhatt}
	\mathit{MHA}(\mat{C}, \mat{H})_i = \sum_{j=1}^{\abs{N}} \alpha_{ij} (\vec{H}_j \mat{W}^{V_t})\\
	\label{eq:mhattalpha}
	\vec{\alpha}_{i} = \softmax{\frac{\vec{C}_i\mat{W}^{Q_t} (\vec{H}\mat{W}^{K_t})^\top}{\sqrt{d}}}
\end{align}

\para{Generation probabilities}
\label{sec:probabilities}
The final representation $\mat{Z}^{(\D)}$ of the last decoder layer $\D$ is used
to compute 
the probability distribution $\vec{P}_{i}\in\interval{0}{1}^{\abs{\Sigma}}$ over all words in the vocabulary $\Sigma$ at time step $i$:
\begin{equation}
	\label{eq:14}
	\vec{P}_i = \softmax{\vec{Z}_i^{(\D)} \mat{E}^\top}
\end{equation}
Note that
$\mat{E}\in \R^{\abs{\Sigma}\times d}$ 
is the same matrix
that is also used to embed node labels and text tokens.

\subsection{Training}
We train Graformer by minimizing the standard negative log-likelihood loss
based on the likelihood estimations described in the previous section. 

\section{Experiments}

\subsection{Datasets}
We evaluate our new architecture on two popular benchmarks
for KG-to-text generation,
AGENDA \citep{koncel-kedziorski-etal-2019-text}
and WebNLG \citep{gardent-etal-2017-webnlg}.
While the latter contains crowd-sourced texts corresponding to subgraphs from various DBPedia categories,
the former was automatically created by applying an information extraction tool \citep{luan-etal-2018-multi}
on a corpus of scientific abstracts \citep{ammar-etal-2018-construction}.
As this process is noisy,
we corrected 7 train instances where an entity name was erroneously split on a special character
and, for the same reason, deleted 1 train instance entirely.
Otherwise, we use the data as is, including the
train/dev/test split.

We list the number of instances per data split,
as well as general statistics about the graphs in \cref{tab:datastats}.
Note that the graphs in WebNLG are human-authored subgraphs of DBpedia
while the graphs in AGENDA were automatically extracted.
This leads to a higher number of disconnected graph components.
Nearly all WebNLG graphs consist of a single component,
i.e., are connected graphs,
whereas for AGENDA this is practically never the case.
We also report statistics that depend on the tokenization (cf.\ \cref{sec:preprocessing})
as factors like the length of target texts
and the percentage of tokens shared verbatim between input graph and target text
largely impact the task difficulty.

\begin{table}[t!]
	\centering
	\small
	\begin{tabular}{lrr}
		\toprule
		& {AGENDA} & {WebNLG} \\
		\midrule
		\#{}instances in train  & 38,719 & 18,102 \\
		\#{}instances in val & 1,000 & 872 \\
		\#{}instances in test & 1,000 & 971 \\
		\midrule
		\#{}relation types & 7 & 373 \\
		avg \#{}entities in KG & 13.4 & 4.0 \\
		\%{} connected graphs & 0.3 & 99.9 \\
		avg \#{}graph components & 8.4 & 1.0 \\
		avg component size & 1.6 & 3.9 \\
		\midrule
		avg \#{}token nodes in graph & 98.0 & 36.0 \\
		avg \#{}tokens in text & 157.9 & 31.5 \\
		avg \%{} text tokens in graph & 42.7 & 56.1 \\
		avg \%{} graph tokens in text  & 48.6 & 49.0 \\
		\midrule
		Vocabulary size $\abs{\Sigma}$ & 24,100 & 2,100 \\
		Character coverage in \%{} & 99.99 & 100.0 \\ 
		\bottomrule
	\end{tabular}
	\caption{Statistics of AGENDA and the dataset from the WebNLG challenge as used in our experiments.
		Upper part: data splits and original KGs. Lower part: token graphs and BPE settings.}
	\label{tab:datastats}
\end{table}

\subsection{Data preprocessing}
\label{sec:preprocessing}

Following previous work on AGENDA \citep{ribeiro20},
we put the paper title into the graph as another entity.
In contrast to \citet{ribeiro20},
we also link every node from a real entity
to every node from the title by \textsc{title2txt} and \textsc{txt2title} edges.
The type information provided by {AGENDA} is,
as usual for KGs,
expressed with one dedicated node per type
and \hasType{} arcs that link entities to their types.
We keep the original pretokenized texts but lowercase the title
as both node labels and target texts are
also lowercased.

For {WebNLG},
we follow previous work \citep{gardent-etal-2017-webnlg}
by replacing underscores in entity names with whitespace
and breaking apart camel-cased relations.
We furthermore follow the evaluation protocol of the original challenge
by converting all characters to lowercased ASCII and
separating all punctuation from alphanumeric characters during tokenization.

For both datasets,
we train a BPE vocabulary using sentencepiece \citep{kudo-richardson-2018-sentencepiece}
on the
train set,
i.e., a concatenation of node labels and target texts.
See \cref{tab:datastats} for vocabulary sizes.
Note that for AGENDA,
only 99.99\%{} of the characters found in the train set
are added to the vocabulary.
This excludes exotic Unicode characters that occur in certain abstracts.

We prepend entity and relation labels with dedicated $\langle E\rangle$ and $\langle R\rangle$ tags.

\subsection{Hyperparameters and training details}
We train Graformer with the Adafactor optimizer \citep{adafactor}
for 40 epochs on AGENDA and 200 epochs on WebNLG.
We report test results for the model yielding the best validation performance measured in corpus-level BLEU \citep{papineni-etal-2002-bleu}.
For model selection,
we decode greedily.
The final results are generated by beam search.
Following \citet{ribeiro20},
we couple beam search with a length penalty \citep{wu16} of $5.0$.
See \cref{app:hyperparam} for more details and a full list of hyperparameters.

\subsection{Epoch curriculum}
We apply a data loading scheme inspired by the bucketing approach of \citet{koncel-kedziorski-etal-2019-text}
and length-based curriculum learning \citep{platanios-etal-2019-competence}:
We sort the train set by target text length and split it into four buckets of two times $40\%$ and two times $10\%$ of the data.
After each training epoch, the buckets are shuffled internally
but their global order stays the same from shorter target texts to longer ones.
This reduces padding during batching as texts of similar lengths stay together
and introduces a mini-curriculum from presumably easier examples (i.e., shorter targets) to more difficult ones
for each epoch.
This enables us to successfully train Graformer \textit{even without a learning rate schedule}.

\section{Results and Discussion}

\subsection{Overall performance}

\cref{tab:agenda} shows the results of our evaluation on AGENDA
in terms of BLEU \citep{papineni-etal-2002-bleu}, METEOR \citep{banerjee-lavie-2005-meteor},
and CHRF++ \citep{popovic-2017-chrf}.
Like the models we compare with,
we report the average and standard deviation of 4 runs with different random seeds.

Our model outperforms previous Transformer-based models
that only consider first-order neighborhoods per encoder layer \citep{koncel-kedziorski-etal-2019-text,an19}.
Compared to the very recent models by \citet{ribeiro20},
Graformer performs very similarly.
Using both a local and a global graph encoder,
\citet{ribeiro20}
combine information from very distant nodes
but at the same time need extra parameters for the second encoder.
Graformer is more efficient and
still matches their best model's BLEU and METEOR scores within a standard deviation.

The results on the test set of seen categories of WebNLG (\cref{tab:webnlg}) look similar.
Graformer outperforms most original challenge participants and more recent work.
While not performing on par with CGE-LW on WebNLG,
Graformer still achieves more than 96\%{} of its performance
while using only about half as many parameters.

\begin{table}[t]
	\centering
	\small
	\def\intercolumn{\hspace{.7em}}
	\begin{tabular}{l@{\intercolumn}l@{\intercolumn}l@{\intercolumn}l@{\intercolumn}l}
		\toprule
		& BLEU & METEOR & CHRF++ & \#{}P \\
		\midrule
		Ours & \meanvar{17.80}{0.31} & \meanvar{22.07}{0.23} & \meanvar{45.43}{0.39} & 36.3 \\
		\midrule
		GT & \meanvar{14.30}{1.01} & \meanvar{18.80}{0.28} & -- & --\\
		GT+RBS & \meanvar{15.1}{0.97} & \meanvar{19.5}{0.29} & -- & --\\
		CGE-LW & \meanvar{18.01}{0.14} & \meanvar{22.34}{0.07} & \meanvar{46.69}{0.17} & 69.8 \\
		\bottomrule
	\end{tabular}
	\caption{Experimental results on AGENDA. GT (Graph Transformer) from \citep{koncel-kedziorski-etal-2019-text}; GT+RBS from \citep{an19}; CGE-LW from \citep{ribeiro20}. Number of parameters in millions.}
	\label{tab:agenda}
\end{table}

\begin{table}[t]
	\centering
	\small
	\def\intercolumn{\hspace{.44em}}
	\begin{tabular}{l@{\intercolumn}l@{\intercolumn}l@{\intercolumn}l@{\intercolumn}l}
		\toprule
		& BLEU & METEOR & CHRF++ & \#{}P\\
		\midrule
		Ours & \meanvar{61.15}{0.22} & \meanvar{43.38}{0.17} & \meanvar{75.43}{0.19} & 5.3 \\
		\midrule
		UPF-FORGe & 40.88 & 40.00 & -- & -- \\
		Melbourne & 54.52 & 41.00 & 70.72 & -- \\
		Adapt & 60.59 & 44.00 & 76.01 & --\\
		\midrule
		Graph Conv. & 55.90 & 39.00 & -- & 4.9 \\
		GTR-LSTM & 58.60 & 40.60 & -- & -- \\
		E2E GRU & 57.20 & 41.00 & -- & -- \\
		\midrule
		CGE-LW-LG & \meanvar{63.69}{0.10} & \meanvar{44.47}{0.12} & \meanvar{76.66}{0.10} & 10.4\\
		\bottomrule
	\end{tabular}
	\caption{Experimental results on the WebNLG test set with seen categories. CGE-LW-LG from \citep{ribeiro20}; Adapt, Melbourne and UPF-FORGe from \citep{gardent-etal-2017-webnlg}; Graph Conv.\ from \citep{marcheggiani-perez-beltrachini-2018-deep}; GTR-LSTM from \citep{trisedya-etal-2018-gtr}; E2E GRU from \citep{castro-ferreira-etal-2019-neural}. Number of parameters in millions.}
	\label{tab:webnlg}
\end{table}

\begin{table}[t]
	\centering
	\small
	\begin{subtable}{\linewidth}
		\centering
		\small
		\begin{tabular}{crrrr}
			\toprule
			$\mu_c$&& BLEU & METEOR & CHRF++ \\
			\midrule
			${} < 1.25$ & Ours & \textbf{15.44} & 20.59 & 43.23 \\
			(213) & CGE-LW & 15.34 & \textbf{20.64} & \textbf{43.56} \\
			\midrule
			${} < 1.5$ & Ours & \textbf{17.45} & 22.03 & 45.67 \\
			(338) & CGE-LW & 17.29 & \textbf{22.32} & \textbf{45.88} \\
			\midrule
			${} < 2.0$ & Ours & 18.94 & 22.86 & 46.49 \\
			(294) & CGE-LW & \textbf{19.46} & \textbf{23.76} & \textbf{47.78} \\
			\midrule
			${}\geq 2.0$ & Ours & \textbf{21.72} & 24.22 & 48.79 \\
			(155) & CGE-LW & 20.97 & \textbf{24.98} & \textbf{49.83} \\
			\bottomrule
		\end{tabular}
		\caption{Average size $\mu_c$ of graph components.}
		\label{tab:compsize}
	\end{subtable}
	
	\vspace{.5em}
	\begin{subtable}{\linewidth}
		\centering
		\small
		\begin{tabular}{crrrr}
			\toprule
			$\mathsf{d}$&& BLEU & METEOR & CHRF++ \\
			\midrule
			1 & Ours & \textbf{16.48} & \textbf{21.16} & 43.94 \\
			(368) & CGE-LW & 16.33 & \textbf{21.16} & \textbf{44.16} \\
			\midrule
			2 & Ours & \textbf{18.46} & 22.70 & 46.85 \\
			(414) & CGE-LW & 18.20 & \textbf{23.14} & \textbf{47.28} \\
			\midrule
			${}\geq 3$ & Ours & 19.44 & 23.17 & 47.29 \\
			(218) & CGE-LW & \textbf{20.32} & \textbf{24.42} & \textbf{49.25} \\
			\bottomrule
		\end{tabular}
		\subcaption{Largest diameter $\mathsf{d}$ across all of a graph's components.}
		\label{tab:diameter}
	\end{subtable}
	\caption{Performance of a single run on the test split of AGENDA w.r.t.\ different input graph properties. The number of data points in each split is indicated in parentheses.}
	\label{tab:analysis}
\end{table}

\begin{figure*}[t]
	\begin{subfigure}{.49\linewidth}
		\centering
		\includegraphics[height=19.5em]{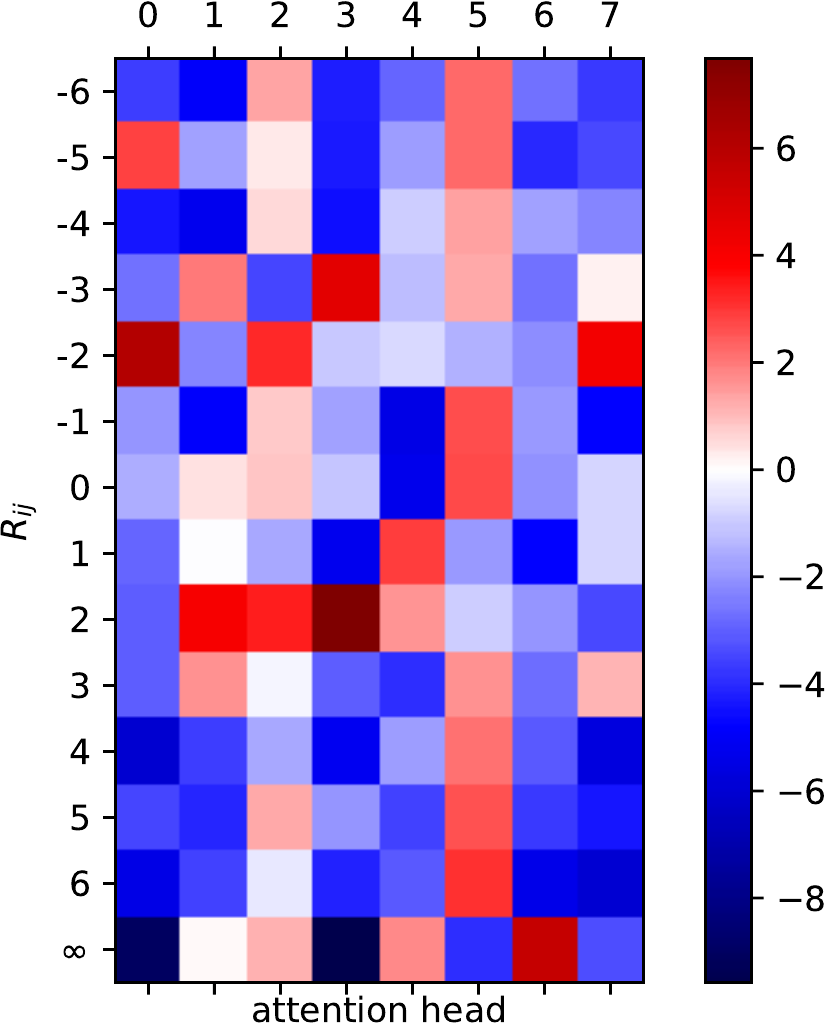}
		\subcaption{AGENDA}
		\label{fig:attention-agenda}
	\end{subfigure}
	\begin{subfigure}{.49\linewidth}
		\centering
		\includegraphics[height=19.5em]{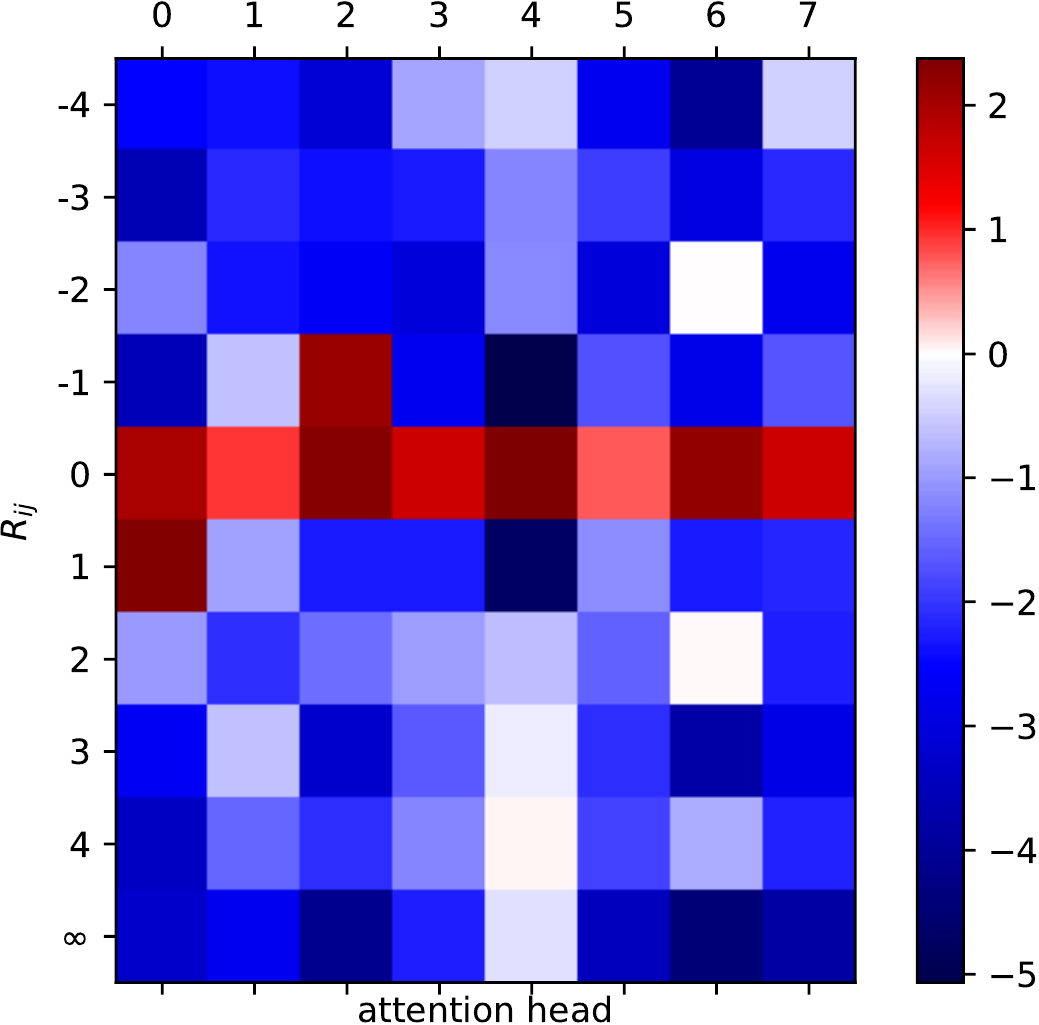}
		\subcaption{WebNLG}
		\label{fig:attention-webnlg}
	\end{subfigure}
	
	\caption{Attention bias $\gamma$ learned by Graformer on the two datasets. $\sameEnt{}_p$ edges are omitted.}
	\label{fig:attention}
\end{figure*}

\subsection{Performance on different types of graphs}

We investigate whether Graformer is more suitable for disconnected graphs
by comparing its performance on different splits of the AGENDA test set according to two graph properties:
\begin{enumerate*}[label=(\roman*)]
	\item the average number of nodes per connected component ($\mu_c$) and
	\item the largest diameter across all of a graph's components ($\mathsf{d}$).
\end{enumerate*}

We can see in \cref{tab:analysis} that
the performance of both Graformer and CGE-LW \citep{ribeiro20}
increases with more graph structure (larger $\mu_c$ and $\mathsf{d}$),
i.e., more information leads to more accurate texts.
Besides, Graformer outperforms CGE-LW on BLEU for graphs with smaller components ($0 < \mu_c < 1.5$)
and smaller diameters ($\mathsf{d}\! <\! 3$).
Although METEOR and CHRF++ scores always favor CGE-LW,
the performance difference is also smaller for cases where BLEU favors Graformer.

We conjecture that
Graformer benefits from its more elaborate global view,
i.e., its ability to distinguish between distant but connected nodes and unreachable ones.
CGE-LW's global encoder cannot make this distinction
because it only sees a fully connected version of the graph.

Curiously, Graformer's BLEU is also better for larger components ($\mu_c \geq 2.0$).
With multiple larger components,
Graformer might also better distinguish nodes that are part of the same component from those that belong to a different one.

Only for $1.5 < \mu_c < 2.0$,
CGE-LW clearly outperforms Graformer in all metrics.
It seems that Graformer is most helpful for extreme cases, i.e.,
when either most components are isolated nodes or when isolated nodes are the exception.

\begin{table}[t]
	\centering
	\small
	\begin{tabular}{lrrr}
		\toprule
		Model & BLEU & METEOR & CHRF++ \\
		\midrule
		Graformer & 18.09 & 22.29 & 45.77 \\
		\midrule
		-length penalty & 17.99 & 22.19 & 45.63 \\
		-beam search & 17.33 & 21.74 & 44.87 \\
		-epoch curriculum & 13.55 & 18.91 & 39.22 \\
		\bottomrule
	\end{tabular}
	\caption{Ablation study for a single run on the test portion of AGENDA.}
	\label{tab:ablation}
\end{table}

\subsection{Ablation study}

In a small ablation study, we examine the impact of beam search, length penalty,
and our new epoch curriculum training.
We find that beam search and length penalty do contribute to the overall performance but
to a relatively small extent.
Training with our new epoch curriculum, however,
proves crucial for good performance.
\citet{platanios-etal-2019-competence} argue that curriculum learning can replace a learning rate schedule,
which is usually essential to train a Transformer model.
Indeed we successfully optimize Graformer without any learning rate schedule, when applying the epoch curriculum.

\section{Learned graph structure}
\label{sec:learned_structure}

We visualize the learned attention bias $\gamma$
for different relative graph positions $R_{ij}$ (cf.\ \cref{sec:self-att}; esp.\ \cref{eq:7})
after training on AGENDA and WebNLG
in \cref{fig:attention}.
The eight attention heads (x-axis)
have learned
different weights for each graph position $R_{ij}$ (y-axis).
Note that AGENDA has more possible $R_{ij}$ values
because $n_\delta = 6$ whereas we set $n_\delta=4$ for WebNLG.

For both datasets,
we notice that one attention head primarily focuses on global information
(5 for AGENDA, 4 for WebNLG).
AGENDA even dedicates head 6 entirely to unreachable nodes, showing the importance of such nodes for this dataset.
In contrast, most WebNLG heads
suppress information from unreachable nodes.

For both datasets,
we also observe that nearer nodes generally receive a high weight (focus on local information):
In \cref{fig:attention-webnlg}, e.g., head 2 concentrates solely on direct incoming edges and head 0 on direct outgoing ones.
Graformer can learn empirically based on its task
where direct neighbors are most important and where they are not, 
showing that the strong bias from graph neural networks is
not
necessary to learn about graph structure.

\section{Conclusion}
We presented
Graformer,
a novel encoder-decoder architecture for graph-to-text generation based on Transformer.
The Graformer encoder uses a novel type of self-attention
for graphs based on shortest path lengths between nodes,
allowing it to detect global patterns
by automatically learning appropriate weights for higher-order neighborhoods.
In our experiments on two popular benchmarks for text generation from knowledge graphs,
Graformer achieved competitive results
while using many fewer parameters than alternative models.

\section*{Acknowledgments}

This work was supported by the
BMBF (first author) as part of the project MLWin (01IS18050),
by the German Research Foundation (second author) as part of the Research Training Group ``Adaptive Preparation of Information from Heterogeneous Sources'' (AIPHES) under the grant No.\ GRK 1994/1,
and by the Bavarian research institute for digital transformation (bidt) through their fellowship program (third author).
We also gratefully acknowledge a Ph.D.\ scholarship
awarded to the first author by the German Academic Scholarship Foundation (Studienstiftung des
deutschen Volkes).

\bibliography{anthology,references}
\bibliographystyle{acl_natbib}

\appendix
\section{Hyperparameter details}
\label{app:hyperparam}

For AGENDA and WebNLG, a minimum and maximum decoding length were set according to the shortest and longest target text in the train set.
\Cref{tab:hparams} lists the hyperparameters used to obtain final results on both datasets.
Input dropout is applied on the word embeddings directly after lookup for node labels and target text tokens
before they are fed into encoder or decoder.
Attention dropout is applied to all attention weights computed during multi-head (self-)attention.

For hyperparameter optimization,
we only train for the first 10 (AGENDA) or 50 (WebNLG) epochs to save time.
We use a combination of manual tuning and a limited number of randomly sampled runs.
For the latter we apply Optuna with default parameters \citep{optuna19,bergstra11} and median pruning, 
i.e., after each epoch of a particular hyperparameter run
we check if the best performance so far is worse
than the median performance of previous runs at the same epoch
and if so, abort.
For hyperparameter tuning, we decode greedily and measure performance in corpus-level BLEU \citep{papineni-etal-2002-bleu}.

\begin{table}[t]
	\centering
	\small
	\begin{tabular}{lrr}
		\toprule
		Hyperparameter & WebNLG & AGENDA\\
		\midrule
		model dimension $d$ & 256 & 400 \\
		\#{} heads & 8 & 8 \\
		\#{} encoder layers $\E$ & 3 & 4 \\
		\#{} decoder layers $\D$ & 3 & 5 \\
		feedforward dimension & 512 & 2000 \\
		attention dropout & 0.3 & 0.1 \\
		dropout & 0.1 & 0.1 \\
		input dropout & 0.0 & 0.1 \\
		text self-attention range $n_t$ & 25 & 50 \\
		graph self-attention range $n_\delta$ & 4 & 6\\
		\sameEnt{} range $n_p$ & 10 & 10 \\
		gradient accumulation & 3 & 2 \\
		gradient clipping & 1.0 & 1.0 \\
		label smoothing & 0.25 & 0.3 \\
		$L_2$ regularizer & $3\cdot10^{-3}$ & $3\cdot10^{-4}$ \\
		batch size & 4 & 8 \\
		\midrule
		\#{} beams & 2 & 2 \\
		length penalty & 5.0 & 5.0 \\
		\bottomrule
	\end{tabular}
    \caption{Hyperparameters used to obtain final experimental results on WebNLG and AGENDA.}
    \label{tab:hparams}
\end{table}

\section{Qualitative examples}
\label{app:qualitative}

\Cref{tab:qualitative_webnlg} shows three example generations from our Graformer model and the CGE-LW system by \citet{ribeiro20}.
Often CGE-LW generations have a high surface overlap with the reference text while Graformer texts fluently express the same content.

\begin{table}[!h]
	\begin{tabularx}{\linewidth}{lX}
		\toprule
		Ref.&julia morgan has designed many significant buildings , including the los angeles herald examiner building .\\
		CGE-LW & julia morgan has designed many significant buildings including the los angeles herald examiner building .\\
		Ours & one of the significant buildings designed by julia morgan is the los angeles herald examiner building .\\
		\midrule
		Ref.&asam pedas is a dish of fish cooked in a sour and hot sauce that comes from indonesia .\\
		CGE-LW & the main ingredients of asam pedas are fish cooked in a sour and hot sauce and comes from indonesia .\\
		Ours & the main ingredients of asam pedas are fish cooked in sour and hot sauce . the dish comes from indonesia .\\
		\midrule
		Ref.&banana is an ingredient in binignit which is a dessert . a cookie is also a dessert .\\
		CGE-LW & banana is an ingredient in binignit , a cookie is also a dessert .\\
		Ours & a cookie is a dessert , as is binignit , which contains banana as one of its ingredients . \\
		\bottomrule
	\end{tabularx}
	\caption{Example references and texts generated by CGE-LW \citep{ribeiro20} and Graformer (marked Ours) for samples from the WebNLG test set. In case of multiple references, only one is shown for brevity.}
	\label{tab:qualitative_webnlg}
\end{table}

\end{document}